# Explainability of Algorithms

**Andrés Páez**

Department of Philosophy and
Center for Research and Formation in
Artificial Intelligence (CinfonIA)
Universidad de los Andes
Bogotá, Colombia

Abstract

The opaqueness of many complex machine learning algorithms is often mentioned as one of the main obstacles to the ethical development of artificial intelligence (AI). But what does it mean for an algorithm to be opaque? Highly complex algorithms such as artificial neural networks process enormous volumes of data in parallel along multiple hidden layers of interconnected nodes, rendering their inner workings epistemically inaccessible to any human being, including their designers and developers; they are 'black boxes' for all their stakeholders. But opaqueness is not always the inevitable result of technical complexity. Sometimes, the way an algorithm works is intentionally hidden from view for proprietary reasons, especially in commercial automated decision systems, creating an entirely different type of opaqueness. In the first part of the chapter, we will examine these two ways of understanding opacity and the ethical implications that stem from each of them. In the second part, we explore the different explanatory methods that have been developed in computer science to overcome an AI system's technical opaqueness. As the analysis shows, explainable AI (XAI) still faces numerous challenges. The last section focuses on the relationship between explainability and trust in AI.

The lack of transparency of many contemporary machine learning (ML) models is often mentioned as one of the main obstacles to the ethical development of artificial intelligence (AI). Most recent AI ethics guidelines include transparency as a fundamental ethical principle. Despite its widespread desirability, there is no standard definition of a transparent algorithm, no common understanding of the properties that an AI model must possess to avoid being catalogued as a 'black box'. Oftentimes, transparency is characterised in terms of explainability, interpretability or understandability, terms that also lack a commonly agreed upon definition. Instead of trying to offer an account of algorithmic transparency or its cognates, this chapter will start at the opposite end of the spectrum, by examining the types and sources of algorithmic opacity. By focusing initially on opacity, it will be easier to assess the epistemic achievements of the methods designed to shed some light on the workings of complex AI algorithms. This approach will also help us understand why opacity is a challenge to the ethical development of AI.

The chapter is organised as follows. In the first section, I establish a distinction between the two main types of algorithmic opacity, one based on legal considerations and the other on epistemic grounds. Both types of opacity have strong ethical implications. Most of the chapter will focus on epistemic opacity. The second section will discuss the current methods used in computer science to overcome epistemic opacity and achieve some degree of algorithmic transparency. One of the conclusions of this section will be that currently there is no clear path towards absolute transparency. In the third section, I explain how these methods provide some degree of understanding of opaque models. The final section focuses on the relation between explainability and trust in automated decisions based on AI systems. The empirical evidence will show that transparency is not always desirable when the goal is to build trust towards AI systems.

## 1. Two Types of Algorithmic Opacity

The expression 'opaque algorithm' can be understood in two very different ways. On the one hand, it may refer to models whose structure, dataset, features, weights and biases, are proprietary, industrial secrets protected by business and copyright laws. These models are not necessarily complex, but their opaque nature stems from the fact that people affected by their decisions are legally impeded to access their data and inner logic. On the other hand, the expression might refer to algorithms that escape human comprehension due to their extreme complexity, making them epistemically inaccessible. There is a well-known trade-off between an algorithm's accuracy and its transparency. Deep neural networks (DNNs) are the paradigmatic example of a highly accurate and highly opaque type of algorithm. In this section, I will discuss both types of opacity and the ethical problems they generate. To facilitate the discussion, I will use the terms *legal opacity* and *epistemic opacity* to refer to the two types of algorithmic opacity.

## 1.1. Legal Opacity

In recent years, both private companies and government agencies have designed and implemented automated decision software about issues that are deeply important in people's lives. These include decisions about access to healthcare, housing, education, subsidies, credit and employment; legal decisions about bail and parole; and different forms of profiling for purposes as diverse as predictive policing and tax fraud detection. We will examine the ethical consequences of having legal statutes that restrict access to the details of these algorithms and protect their opaque nature.

Perhaps the best-known example of a legally opaque algorithm is COMPAS (Correctional Offender Management Profiling for Alternative Sanctions), an algorithm mostly used in the American legal system to assess the risk of recidivism of people at different stages of a legal process. The algorithm was created and is owned by a private company called Northpointe (now Equivant). The algorithm generates risk scales for general and violent recidivism, which are used in parole decisions, and risk scales for pretrial misconduct such as committing new felonies or failing to appear in court. The secret nature of the algorithm became a legal point of dispute in *State v. Loomis* (2016), a case in which the defendant, Eric Loomis, claimed that using the risk assessment provided by COMPAS in his sentencing decision violated his right to due process because it violated his right to be sentenced on accurate information (*Loomis*, p. 757). The algorithm's legal opacity prevented him from challenging its scientific validity and accuracy. The Wisconsin Supreme Court dismissed his arguments and affirmed the lower court's decision.

The question *Loomis* did not address regarded the risks involved in using opaque algorithms in legal decisions. COMPAS and similar proprietary risk assessment algorithms have been subject to criticism for reinforcing existing inequalities, violating

Equal Protection Rights on the basis of race, and disguising overt discrimination based on demographics and socio-economic status. COMPAS has also been accused of being less accurate when assessing Black defendants, although this claim has been disputed by the company. It has also been found to fail group fairness metrics. In general terms, actuarial and algorithmic risk assessment tools suffer from poor data quality and uncertainty problems, which makes their implementation a factually irresponsible policy.[1]

There is no easy remedy for legal opacity. Deep changes in business and copyright laws are unlikely, and many sectors seem keen on adopting proprietary algorithms. From my perspective, the main risk of the widespread adoption of these tools is that they will increasingly replace human judgement as people end up uncritically accepting the scores and recommendations generated by the algorithms. Empirical research supports this conclusion. Results in behavioural economics and social psychology show that it is psychologically difficult and rare to override algorithmic recommendations and that algorithmic outputs tend to act as anchors for agent's decisions, resulting in a displacement of discretion. There is no guarantee that gaining access to the details of the proprietary algorithms would counter that tendency, but at least there would be more accountability as the weights given to each variable become known. Decision-makers would become responsible for knowingly basing their decisions on the potentially biased or discriminatory variables of the algorithms.

## 1.2. Epistemic Opacity

The second kind of algorithmic opacity is not the result of anyone's actions or decisions. It is, rather, a consequence of the way that complex ML algorithms work. Not all AI models are opaque. Some use simple architectures like decision trees that do not require any technical knowledge to be understood. But the workings of the most powerful algorithms with the strongest predictive power are beyond human comprehension. Consider the case of DNNs, which are the most common example of an opaque algorithm.

DNNs are designed to identify patterns and correlations in vast amounts of data, many of which are not symbolic and therefore incomprehensible to humans. The network then uses these patterns to make the predictions and classifications for which it has been trained. Inputs pass through multiple hidden layers of interconnected nodes, or 'neurons', each of which transforms them in different ways before passing them along to the next layer. These transformations often involve nonlinear operations. These nonlinear transformations, combined with the interactions between the neurons across different layers, allow the neural network to model complex, high-dimensional decision boundaries. Even if we could see all the weights and biases[2] in the network (the parameters that the model learned during training), it is not clear how to interpret them in the context of the original input features. Each input of the network goes through a series of complex, intertwined transformations that make it impossible to understand how the inputs relate to the output. The nonlinear nature of the activation functions compounds this complexity. In a linear system, the effect of each input on the output can be considered independently of the others, but in a nonlinear system, the effect of changing an input depends on the values of all the other inputs. This makes it epistemically impossible to understand how each input influences the output.

The second problem is that there is no way of verifying which parameters are being used in the hidden layers of the DNN. Deep models often have an extremely large number of optima of similar predictive accuracy. In other words, there are many possible settings of the weights and biases of a model that will result in equally good outputs, and there is no way of identifying the settings of a particular model. This is known as the *model identifiability problem*: 'A model is said to be identifiable if a sufficiently large training set can rule out all but one setting of the model's parameters. Models with latent variables are often not identifiable because we can obtain equivalent models by exchanging latent variables with each other' (Goodfellow et al. 2016, p. 284). Without identifying which model is being used, it is impossible to 'explain' the model's prediction. There *is*, of course, a true description of the model, but it is inaccessible to human subjects.

Epistemic opacity is a problematic feature of algorithms for several reasons. From an ethical perspective, it raises many of the same questions as legal opacity regarding hidden discrimination and the violation of human rights. On the technical side, it is an obstacle to developers who want to be able to debug a system, improve model performance, and detect and resolve bias and other potential risks. From a regulatory perspective, the General Data Protection Regulation (GDPR) and more recent regulations in different countries require that the logic behind automated systems whose outcomes have a significant effect on people's lives should be made available to end-users. It is still an open question whether the technological sector will be able to comply with this legal requirement. Epistemic opacity is also believed to be an obstacle in building user trust in a model, a claim that I will discuss in Section 4. For these, and many other reasons, different methods have been devised to eliminate, or at least ameliorate epistemic opacity. We now turn to these methods.

## 2. XAI and the Efforts to Restore Transparency

Explainable AI (XAI) is a research program in computer science that aims to bring some light into epistemically opaque ML models. One option is to try to avoid opacity from the outset, by solving predictive problems using ML models that are easier to understand, while maintaining a high level of accuracy. However, this only works for certain types of tasks but is unfeasible in others. The most common approaches in XAI are either to try to provide an explanation of a *singular* prediction of an opaque model or to try to explain the workings of the model *globally*, by providing a general idea of its functioning and capabilities through a simpler model. The former approach is based on so-called local *post-hoc* interpretability methods, which include counterfactual probes and different types of perturbation-based methods such as Local Interpretable Model-Agnostic Explanations (LIME), Gradient-weighted Class Activation Mapping (Grad-CAM), Shapely Additive Explanations (SHAP), Testing with Concept Activation Vectors (TCAV), among many others.[3] The common principle of all these methods is to make systematic changes to the input to determine which features are most responsible for a given output. The global approach is based on proxy, interpretative, or surrogate models. These are very simple models that replicate to a certain extent the way the opaque model works. The most widely used classes of surrogate models are linear or gradient-based approximations, decision rules, and decision trees. Neither of these approaches to XAI is ideal, as I aim to show in this section.

During a long time, local *post-hoc* interpretability methods were seen – at least within the community of AI developers – as the most promising approach to open the black box of AI. Recently, however, they have become the subject of much criticism due to their intrinsic limitations and weaknesses, and because of the inscrutability of the resulting 'explanations' for non-expert end-users and stakeholders. Perhaps the most damning problem for local methods is that they perform poorly on diverse robustness metrics (Kindermans et al. 2019). Ideally, very similar inputs should not generate substantially different explanations. But simple transformations of the input, or repeating the sampling process, can generate different explanations. Also, adding a constant shift to the input data, which is a simple and common pre-processing step with no effect on the performance of the model, causes numerous interpretation methods to make incorrect attributions. LIME and SHAP are vulnerable to adversarial attacks, such as perturbations that produce perceptively indistinguishable inputs that are assigned the same predicted label yet result in very different interpretations.

Another limitation of local *post-hoc* interpretability methods such as heat and saliency maps, which highlight in red the parts of an image most responsible for the outcome, is that they lack precision. Rajpurkar et al. (2017), for example, propose a heatmap method in medical AI. The method highlights the areas of a patient's X-ray deemed most important for the diagnosis of pneumonia by a DNN. However, Ghassemi et al. (2021) argue that even the hottest parts of a map contain both useful and useless information (from the perspective of a human expert), and simply localising the region does not reveal exactly what it was in that area that the model considered important:

> The clinician cannot know if the model appropriately established that the presence of an airspace opacity was important in the decision, if the shapes of the heart border or left pulmonary artery were the deciding factor, or if the model had relied on an inhuman feature, such as a particular pixel value or texture that might have more to do with the image acquisition process than the underlying disease (Ghassemi et al. 2021, p. e746).

Furthermore, the information provided in the hot area must be interpreted, thereby opening the door to the clinician's previous beliefs and the risk of confirmation bias. The explanation also lacks any sort of justification of why that particular area was more relevant than others because there is no causal knowledge supporting the explanation. Finally, automation bias can lead to an overestimation of the ML system's performance.

Counterfactual methods are not immune to the robustness problem. Like perturbation-based methods, counterfactual methods can be manipulated and may converge towards drastically different explanations under small perturbations (Slack et al. 2021). Counterfactual probes also critically depend on closeness metrics but there is no principled way to decide which metric to use in any given case. And like saliency-based methods, the lack of causal grounding can deliver suboptimal or even erroneous explanations to decision-makers.

This is just a small sample of the problems faced by local explanations. Their fragility, imprecision, and lack of generality are grave enough to recommend at least combining them with global methods to achieve a better understanding of ML systems. To be sure, global explanation methods are not a panacea. To create a manageable linear model, expert knowledge is required to select the features that will be included. Only features that exceed a certain threshold of correlation between the feature and the target should be used, but there is always the risk that some features might not show an individual correlation, and/or that their contribution only becomes clear in combination with other features. The advantage of surrogate linear models is that they are widely used in the natural and social sciences, including medicine, which makes them a familiar and accepted tool for many of their intended users. Decision trees are used in cases where the relation between features and targets is nonlinear or where features interact with each other. They can also be expressed as decision rules. However, their step-by-step nature is not very efficient. They are also very sensitive to small changes in the training dataset or to a change in feature choice: a change in a split high up the tree will affect the entire tree. Furthermore, many global methods that attempt to preserve the performance of the original model end up generating grey boxes. And when the surrogate models are easy to understand, they incur in overfitting and deteriorated accuracy compared to the teacher model. Of course, one can argue that the main goal of a surrogate model is not to achieve a similar level of performance to the original model, but rather to help users obtain a basic understanding of how it works. Simple explanations might help users understand the capabilities and limitations of the model and adjust their confidence levels accordingly. Simple decision trees, rule lists, example-based methods, and even dialogical explanations will perform much better in this kind of task.

But what exactly does it mean to say that an XAI method makes a model *understandable*? How can we characterise the common goal of all these methods? We now turn to this question.

## 3. Explainability and Understanding

At first glance, understanding a specific decision of a system – through a local *post-hoc* interpretation method – and understanding an ML model globally – using a decision tree, for example – are two mental states that demand different accounts of understanding. The former seems to correspond to a form of *understanding-why*, while the latter to a form of *objectual understanding*. Both types of understanding have been widely discussed in epistemology and the philosophy of science. Let us examine each one in turn.

The characterisation of objectual understanding in the epistemological literature fits well with the purpose of surrogate models. Zagzebski, for example, argues that understanding 'involves grasping relations of parts to other parts and perhaps the relation of parts to a whole' (2009, p. 144). The kinds of relations she has in mind can be spatial, temporal or causal. Her account assumes that objectual understanding requires being able to identify the various parts of the object, describe their functional interdependence, and use that information to make useful inferences. This is precisely what a surrogate model offers, either through association rules or directly visible on a decision tree. It aims at providing a simplified version of the original model by providing a view of its features and structure, and of the possible functional interactions between the features. The level of complexity of surrogate models is directed entirely by pragmatic considerations. Once the user has understood the functional relations of interest for his epistemic needs, it can be said that the model has become transparent to him. Transparency is a success concept that depends on being able to grasp the functional structure of the model through the surrogate model.

The epistemological accounts of understanding-why fit well with local *post-hoc* interpretability methods. Understanding why *p* is not equivalent to simply *knowing* why *p*. Knowing that an image recognition system correctly classified an image as a dog because it was presented with an image of a dog is clearly not enough to understand the decision. The person must be able to answer a wide range of counterfactual questions of the type what-if-things-had-been-different. What if the ears of the dog were not visible? What if the light had been weaker? What if the input were the mirror image of the original? Local interpretation methods allow users to visualise variations of the input in ways that provide answers to these questions.

Many authors have argued that understanding *in general* requires the ability to envision different configurations of the parts of an object and infer its resultant states. In other words, they claim that understanding requires the ability to think counterfactually. Understanding the possible configuration of an object is the way to understand why it emerged and how it functions. But appearances to the contrary, local *post-hoc* interpretability methods do not provide genuine counterfactual understanding. By tinkering with the input, users can only establish piecemeal correlations that cannot be generalised in any obvious way. What prevents users from engaging in genuine counterfactual reasoning is the lack of structural and functional information about the workings of the system, that is, the lack of objectual understanding. Surrogate models provide general rules that have been mined from the data or extracted directly from the model, thus providing the underlying functional scaffolding required to reason counterfactually. One can follow one branch or another of a decision tree, and each path will be a counterfactual case that will be entirely determined by the static functional structure depicted in the tree. Understanding-why thus always requires some degree of objectual understanding.

Karimi et al. (2021) offer an argument in a similar direction but from a more practical perspective. They focus on the problem of algorithmic recourse. When a person has been affected by an unfavourable automated decision (e.g. a rejected housing application), counterfactual explanations can be used to suggest actions that a person can take to achieve a favourable decision from an ML system. The authors show that such counterfactuals 'do not translate to an *optimal* or *feasible* set of actions that would favorably change the prediction of *h* if acted upon. This shortcoming is primarily due to the lack of consideration of causal relations governing the world' (p. 359). The missing causal information is part of the theoretical knowledge included in the objectual understanding of the system. To be sure, surrogate models by themselves do not provide the missing causal information, but they at least provide human-interpretable functional correlations that can be empirically investigated and validated. They can be seen as an initial step towards obtaining the causal knowledge required to design actionable decision systems.

In sum, local *post-hoc* interpretability methods by themselves cannot provide understanding of a model. They only provide the causes of specific predictions of the model, establishing thereby the interconnections between features and outputs. However, only an agent's ability to reason counterfactually about the model, only her ability to use it and manipulate it, can be taken as evidence that she has understood it, that it has become transparent to her. Surrogate models are perhaps the best epistemic tool available for a wide variety of stakeholders to master the workings of opaque target models.

## 4. Explainability and Trust

The final question I will address is whether acquiring the ability to use and manipulate a model results in an increased level of trust in its predictions. There is a widespread belief that transparency and explainability build trust in AI. The *UNESCO Recommendation on the Ethics of Artificial Intelligence*, for example, devotes a chapter to transparency and explainability, and states that they aim 'at providing appropriate information to the respective addressees to enable their understanding and foster trust' (III, §39). In a similar vein, the ethical framework AI4People set forth by Floridi et al. (2018) states that 'it is especially important that AI be explicable, as explicability is a critical tool to build public trust in, and understanding of, the technology' (p. 701). But despite the intuitive appeal of the idea that explainability fosters trust, both the available evidence and the nature of trust itself complicate this simple picture.

Before examining the connection between explainability and trust, it is important to examine the nature of trust itself. In human–human interactions, honesty, competence, and value similarity are essential to establish both *cognitive* and *emotional* trust. Cognitive trust is based on good rational reasons, on one's acquaintance with the trustee, and on evidence about his or her reliability. Emotional trust is based on the positive feelings generated by our interactions with others. It is highly contextual and depends on social and cultural features not easily encoded. The two types of trust can exist independently from each other. People often feel they can trust someone they barely know, based perhaps on social cues; and people trust individuals they know to be reliable but towards whom they have no emotional attachment whatsoever.

Honesty, competence, and value similarity can only be ascribed to others by attributing to them the adequate intentions and beliefs from which these traits can be inferred. In the case of AI, honesty and value similarity are largely irrelevant. For the most part, trust in AI boils down to competence and reliability, that is, to cognitive trust.

In the context of AI, user trust can be understood as 'the extent to which the trustee believes that an automated system will behave as expected' (Papenmeier et al. 2022). However, honesty and value similarity are still relevant in the analysis because they are probably the source of people's preferences for many human-based decisions, even when they are less accurate and reliable than automated ones. For example, there is evidence that patients are reluctant to use health care provided by medical AI even when it outperforms human doctors (Longoni et al. 2019). This behaviour is called 'algorithmic aversion.'

Part of the reason for this preference is that humans judge humans and machines differently. In a recent study, Hidalgo and collaborators compared people's reactions to a wide variety of actions performed by humans and machines. They concluded that, in general, 'humans are judged by their intentions, while machines are judged by their outcomes' (Hidalgo et al. 2021, p. 139). An older meta-analysis of factors affecting trust in human–robot interaction also revealed that 'robot characteristics, and in particular, performance-based factors, are the largest current influence on perceived trust in HRI [Human-Robot Interaction]' (Hancock et al. 2011, p. 523). In the field of medical AI, Hatherley (2020) argues that it is a mistake to use the categories considered to be relevant for interpersonal trust to interactions between humans and medical AI. These systems can be relied upon, but they do not appear to be the appropriate objects of trust. In a similar vein, Ferrario et al. (2020) argue that a medical doctor's reliance on an AI system does not require monitoring it for properties that only humans can have. The meta-analysis performed by Hancock et al. (2011) also found that factors related to human attitudes towards robots had a very small role in trust building.

It seems, therefore, that reliability is the key factor in trust-building towards machines, and that often it is not enough, as the evidence from medical AI seems to indicate. The question, then, is whether explainability can be brought into the picture to complement reliability as a source of trust. Here the evidence is mixed. Some studies seem to indicate that is effective. Shin (2021) conducted a study with 350 individuals who were users of algorithmic news services. The results indicate that transparency and explainability positively impact user trust. In clinical decision support, Wysocki et al. (2023) reported that transparency and explicability were effective in building trust among physicians, although they accentuated confirmation bias and model over-reliance.

Despite these positive results, the evidence for the ineffectiveness of explainability outweighs the evidence for it, even in the same sectors and in similar contexts. Papenmeier et al. (2019) found that high-fidelity explanations actually *decreased* user's trust in high accuracy classification algorithms used in social media. Schmidt et al. (2020) also reported that higher transparency in a text classification task can actually have a negative impact on trust. Furthermore, 'this effect occurs predominantly for cases in which the ML system's predictions are correct, showing that improvident use of transparency within assistive AI tools can in fact impair human performance' (p. 261). These are just some of many examples where empirical research has found no support for the positive relation between explicability and trust.

This literature review indicates that there is a highly contentious debate around the effectiveness of explainability in building trust. Implicit in this discussion is the premise that trust in AI is a desirable goal. Distrust in useful, explainable, and highly reliable AI decision systems seems irrational, even unethical when it prevents people from receiving its benefits without any significant risk. However, there are cautionary voices. Ghassemi et al. (2021) warn about the danger of using unreliable or superficial explanations in healthcare applications that might generate false hope in AI and lead to automation bias. The authors go as far as cautioning against having explainability be a requirement for clinically deployed models. Lakkaraju and Bastani (2020) also warn against XAI methods that only optimise fidelity – that is, that only ensure that the explanations accurately mimic the predictions of black box model – because high fidelity can be reached even if explanations use entirely different features compared to the black box. High fidelity explanations 'can actually mislead the decision maker into trusting a problematic black box' (Lakkaraju and Bastani 2020, p. 79). Even when XAI methods improve self-reports of trust and understanding, there is evidence that these self-reports do not translate into improved performance in tasks using AI support (Papenmeier et al. 2022; Kandul et al. 2023).

5. Concluding Remarks

It is often taken for granted that designing explainable ML systems should be a goal of AI. One of the main reasons offered for that desideratum is that it is desirable for people to trust these systems. The analysis presented in this chapter shows not only that the connection between explainability and trust is far from obvious, but also that currently the degree of transparency to which we can aspire is extremely limited. This has led some researchers to adopt a sceptical stance towards understanding in ML. Humphreys, for example, argues that 'we must abandon the insistence on epistemic transparency for computational science' (2004, p. 150). However, there are epistemic, ethical and legal reasons – that is normative reasons – for XAI. In consequence, the discussion about the possibility of XAI should not be limited to its technical feasibility. It is also a philosophical discussion about the transparency of a decision-making technology that greatly impacts human lives.

## Further Reading

## Notes

[1] Another well-known, problematic example of legal opacity is SyRI (System Risk Indication), an algorithm developed by the Dutch government in 2014 as a welfare and tax fraud risk-scoring instrument.

[2] Weights determine the strength of the connections between neurons. Biases are constants associated with each neuron that serve as a form of threshold, allowing neurons to activate even when the weighted sum of their inputs is not sufficient on its own.

[3] See Ivanovs et al. (2021) for a survey of local XAI methods.